\documentclass[journal]{IEEEtran}
\usepackage{url}
\usepackage{times}
\usepackage{epsfig}
\usepackage{graphicx}
\usepackage{amsmath}
\usepackage{amssymb}
\usepackage{multirow}
\usepackage{booktabs}
\usepackage{setspace}
\usepackage{color}
\usepackage{algorithm}
\usepackage{algpseudocode}
\usepackage{algorithmicx}
\usepackage{amsmath}
\usepackage{threeparttable}
\usepackage{cite}
\usepackage[colorlinks=true,linkcolor=red,citecolor=blue]{hyperref}
\newcommand{\tabincell}[2]{\begin{tabular}{@{}#1@{}}#2\end{tabular}}
\ifCLASSINFOpdf
\else
\fi
\begin{document}
\title{Fingerprint Presentation Attack Detector Using Global-Local Model}
\author{
Haozhe Liu, Wentian Zhang, Feng Liu$^*$, Haoqian Wu, Linlin Shen
\thanks{Manuscript received 4 October 2020; revised 13 March 2021; accepted 15 May 2021. Date of publication 16 June 2021; date of current version 18 Feb 2024. This work was supported in part by the National Natural Science Foundation of China under Grant 62076163 and Grant 91959108; in part by the Shenzhen Fundamental Research Fund under Grant JCYJ20190808163401646 and Grant JCYJ20180305125822769; and in part by the Tencent “Rhinoceros Birds”—Scientific Research Foundation for Young Teachers of Shenzhen University. This article was recommended by Associate Editor P. P. Angelov. \textit{(Corresponding author: Feng Liu.)}}
\thanks{The authors are with the Computer Vision Institute, College of Computer Science and Software Engineering, SZU Branch, Shenzhen Institute of Artificial Intelligence and Robotics for Society, the National Engineering Laboratory for Big Data System Computing Technology, and the Guangdong Key Laboratory of Intelligent Information Processing, Shenzhen University, Shenzhen 518060, China (e-mail: feng.liu@szu.edu.cn).}
\thanks{Digital Object Identifier 10.1109/TCYB.2021.3081764}
}
\IEEEpubid{\begin{tabular}[l]{c@{}} 2168-2267 ~\copyright~2021 IEEE Person use is permitted, but republication/redistribution requires IEEE permission.
\\
See \url{http://www.ieee.org/publications_standards/publications/rights/index.html} for more information.
\end{tabular}
}
\markboth{IEEE TRANSACTIONS ON CYBERNETICS,~Vol.~52, No.~11, November~2022}
{Liu \MakeLowercase{\textit{et al.}}: Fingerprint PA Detector using Global-Local Model}
\maketitle

\begin{abstract}
The vulnerability of automated fingerprint recognition systems (AFRSs) to presentation attacks (PAs) promotes the vigorous development of PA detection (PAD) technology. However, PAD methods have been limited by information loss and poor generalization ability, resulting in new PA materials and fingerprint sensors. This paper thus proposes a global-local model-based PAD (RTK-PAD) method to overcome those limitations to some extent. The proposed method consists of three modules, called: 1) the global module; 2) the local module; and 3) the rethinking module. By adopting the cut-out-based global module, a global spoofness score predicted from nonlocal features of the entire fingerprint images can be achieved. While by using the texture in-painting-based local module, a local spoofness score predicted from fingerprint patches is obtained. The two modules are not independent but connected through our proposed rethinking module by localizing two discriminative patches for the local module based on the global spoofness score. Finally, the fusion spoofness score by averaging the global and local spoofness scores is used for PAD. Our experimental results evaluated on LivDet 2017 show that the proposed RTK-PAD can achieve an average classification error (ACE) of 2.28\% and a true detection rate (TDR) of 91.19\% when the false detection rate (FDR) equals 1.0\%, which significantly outperformed the state-of-the-art methods by $\sim$10\% in terms of TDR (91.19\% versus 80.74\%).
\end{abstract}

\section{Introduction}
\label{sec:Intro}

\IEEEPARstart{T}{he} increasing intelligent devices and the held online payment tools raise an emerging requirement for Automated Fingerprint Recognition Systems (AFRSs).
\cite{cappelli2010minutia,ghafoor2019fingerprint,das2018recent}. The system is designed to verify user authentication \cite{chugh2019oct} with high-level precision and robustness.  
However, current AFRSs are vulnerable to the presentation attacks (PAs) \cite{goicoechea2016evaluation} based on some inexpensive substances like silica gel\cite{liu2019high,sousedik2014presentation}.
, which raises wide concerns about the security of AFRSs. For instance, a Brazilian doctor was arrested for fooling the biometric attendance system at a hospital using spoof fingers made from silicone \cite{marasco2015survey}. In another case, a German group announced that an iPhone equipped with the Touch ID could be fooled by using a sheet of latex or wood glue hosting the fingerprint ridges \cite{marasco2015survey}. Besides those examples mentioned above, many attacks go unreported \cite{marasco2015survey}.

Regarding the system security problem, numerous presentation attack detection (PAD) \cite{marasco2015survey,chugh2019oct} approaches are designed, which can be grouped into two categories, including hardware and software-based methods.
In the case of hardware-based methods, special types of sensors are employed to capture the liveness characteristics (e.g. blood flow \cite{reddy2008new}, pores \cite{zhao2010adaptive} and depth information \cite{liu2019high, liu2020zero}) from fingertips.
For example, Nixon et al. \cite{nixon2004novel} proposed employing the spectral feature of the fingertip, which can be extracted from a spectroscopy-based device, to distinguish between PA and real fingerprint.
Zhao et al. \cite{zhao2010adaptive} extracted the pore from a high-resolution fingerprints. As the pore is hard to re-conducted via artificial attacks, this method can be then utilized for PAD. 
Liu et al. \cite{liu2019high} proposed using Optical Coherence Technology (OCT) to obtain the depth-double-peak feature and sub-single-peak feature, which can achieves 100\% accuracy on their in-hourse dataset for PAD.
These methods require additional hardware in conjunction with the biometric sensor, which makes the device expensive and limits the application field.

Unlike hardware-based methods, software-based solutions work with any commodity fingerprint readers \cite{chugh2019fingerprint}.
Typical software-based methods depend on hand-crafted features, such as physiological features (e.g. perspiration \cite{jia2007new} and ridge distortion \cite{antonelli2006fake}) and texture-based features \cite{lee2009fake, nikam2008fingerprint}.
Antonelli et al. \cite{antonelli2006fake} requires users to move the pressing finger such that the system can detect the PA based on ridge distortion.
Nikam et al. \cite{nikam2008fingerprint} proposed to learn the textural characteristic, which can also achieve a satisfactory result(from 94.35\% to 98.12\%).
However, the aforementioned PAD methods are sensitive to noise and have poor generalization performance.
\IEEEpubidadjcol

Consequently, new methods for fingerprint PAD have been proposed, which are based on convolution neural network (CNN).
Since the performance is reported by the famous publicly available LivDet databases, CNN-based methods outperform the solutions using hand-crafted features by a wide margin \cite{chugh2019fingerprint, menotti2015deep, nogueira2016fingerprint, chugh2018fingerprint}.
Typically, Nogueira et al. \cite{nogueira2016fingerprint} proposed a transfer learning-based method, where deep CNNs, which were originally designed for object recognition and pre-trained on the ImageNet database, were fine-tuned on fingerprint images to differentiate between live and spoof fingerprints.
Chugh et al. \cite{chugh2018fingerprint} proposed a deep convolutional neural-network-based approach using local patches centered and aligned using fingerprint minutiae.
Both of the aforementioned methods improve the performance of PAD, however, there are still some unsolved problems.

\begin{figure}
  \centering
  \includegraphics[width=.47\textwidth]{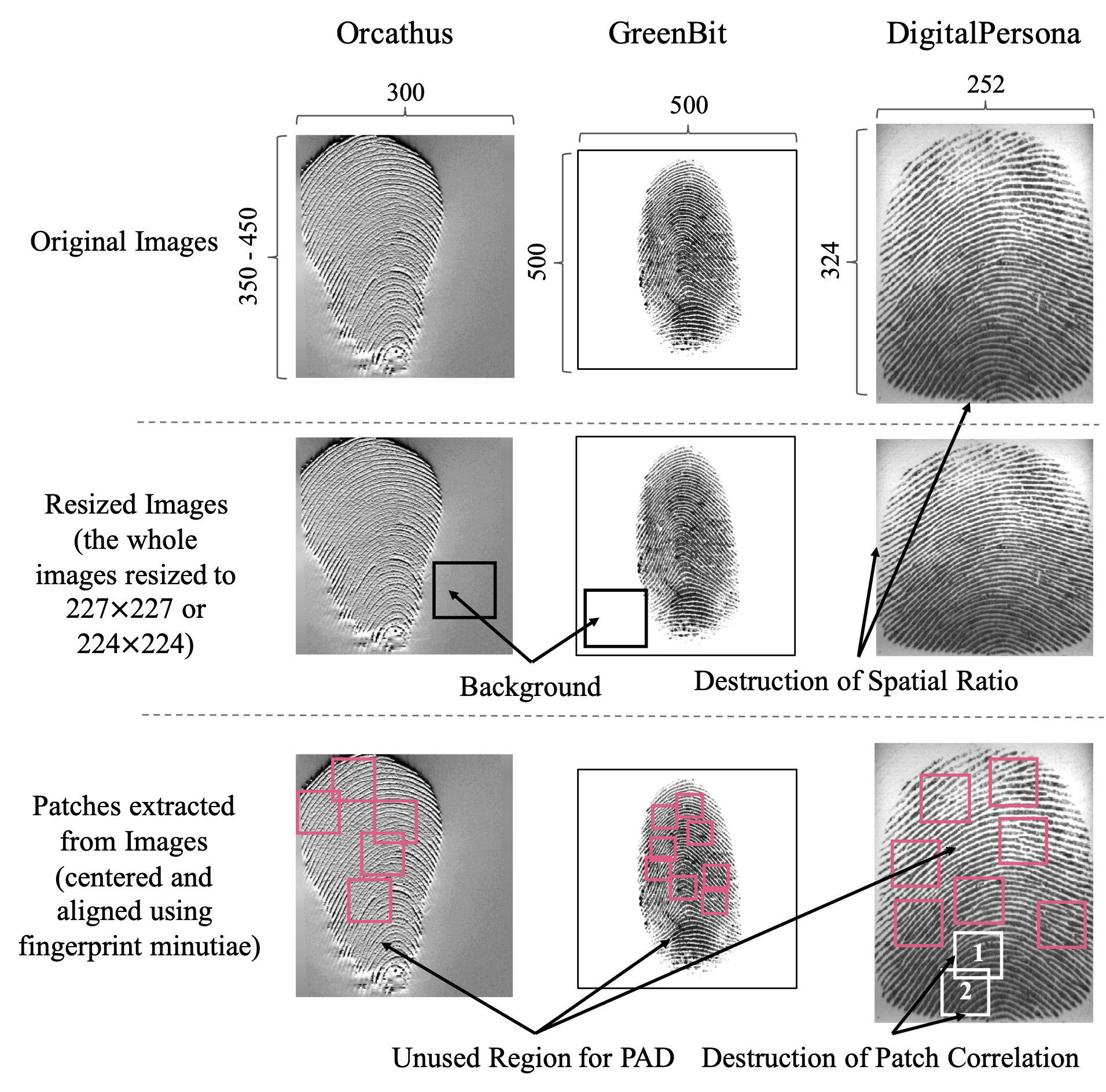}
  \caption{Example of inputs for the fingerprint PAD methods using entire images or patches. The first row shows images from LivDet 2017 \cite{mura2018livdet}. The second row shows the resized images as input for the method \cite{nogueira2016fingerprint}. The third row shows the patches centered and aligned using fingerprint minutiae, which are the inputs of the patch-based method \cite{chugh2018fingerprint}. Columns (left to right): fingerprint images derived from sensors called Orcathus, GreenBit, and DigitalPersona respectively.  }
  \label{fig:input}
\end{figure}

In the method proposed by Nogueira et al. \cite{nogueira2016fingerprint}, the input of deep CNNs is the entire fingerprint image resized to 227 $\times$ 227 pixels for VGG \cite{simonyan2014very} and 224 $\times$ 224 for AlexNet \cite{krizhevsky2012imagenet}.
As shown in Fig. \ref{fig:input}, there are two disadvantages of this method.
Directly resizing the images destroys the spatial ratio of the images and leads to a significant loss of discriminatory information.
In addition, as shown in the second column of Fig. \ref{fig:input}, fingerprint images from some sensors (e.g. GreenBit) have a large blank area surrounding the region of interest (ROI).
This indicates the method using entire images as input is disturbed by the invalid area, which guides the model to learn global features but ignores local features.
In the method \cite{chugh2018fingerprint}, deep CNNs learn from the aligned local patch for PAD. Although this method solved the above-mentioned problems, it introduces new information loss since only the patches around minutiae are used for PAD. The information of other regions in ROI is not considered.
Moreover, the correlation between patches is not used in this method. As the white boxes shown in Fig. \ref{fig:input}, there is an explicit positional correlation between 'box 1' and 'box 2', i.e. 'box 1' should be placed in the upper right corner of ‘box 2’, but deep CNNs miss this geometric information when patches are given as input.

In order to solve these aforementioned problems, this paper proposed a novel PAD method based on the global-local model.
Two different types of input (i.e. the entire images and patches) are given for our proposed method and a rethinking strategy is used to fuse the information from different inputs.
The rethinking strategy mimics the process of humans observing objects, i.e., the human visual cortex is enhanced by top-down stimuli and non-relevant neurons will be suppressed in feedback loops when searching for objects.
Specifically, in rethinking strategy, the proposed model infers twice. The first inference obtains the activation map and gradient of the model, and the second inference uses the information provided by the first inference to get more accurate results.
In the proposed method, the entire image-based model is used for the first inference, and the patch-based model is used for the second inference.
Since different types of input are used in this rethinking strategy, the proposed method not only avoids the conflict of the inputs but also improves the performance of PAD.

This paper is organized as follows: In Section \ref{sec:RelatedWork}, we give a brief overview of the state-of-the-art techniques in PAD and the rethinking strategy used in deep CNNs. Section \ref{sec:method} introduces the proposed PAD method in detail. Then, Section \ref{sec:experiment} shows the empirical evaluation of the proposed method in LivDet and comparisons with the state-of-the-art PAD methods. Finally, we conclude the paper in Section \ref{sec:conclusion}.
\section{Related Work}
\label{sec:RelatedWork}
Since the framework of the proposed PAD method is different from current PAD methods, we reviewed not only the literature on current PAD methods but also the rethinking strategy used in the computer vision field in this section.

Fingerprint PAD is an active research field \cite{marasco2015survey,chugh2019oct,chugh2018fingerprint,nogueira2016fingerprint,chugh2019fingerprint,chugh2019generalization}.
In order to promote technical advances in this field, LivDet competitions have been held since 2009 \cite{ghiani2017review}.
The best result reported in LivDet 2015 \cite{marcialislivdet} is the transfer learning-based method proposed by Nogueira et al. \cite{nogueira2016fingerprint} with an overall accuracy of 95.5\%, where the pre-trained CNN model is used for fingerprint PAD.
In LivDet 2017 \cite{mura2018livdet}, the state-of-the-art detector is proposed by Chugh et al. \cite{chugh2018fingerprint}, which achieves 4.56\% Average Class Error (ACE).
Different from the method \cite{nogueira2016fingerprint} using the entire images as input, this detector takes the patches aligned and centered using minutiae features as the input.
Although CNN-based methods have made some breakthroughs in the research of PAD, some studies \cite{rattani2015open, chugh2019fingerprint, nogueira2016fingerprint} have shown that these CNN-based methods can be attacked by the unpredictable PA (not seen during the training stage of methods).
To generalize the method's effectiveness across different spoof materials, called cross-material performance \cite{chugh2018fingerprint}, two categories of method are proposed, including one-class based \cite{rattani2015open, engelsma2019generalizing, ding2016ensemble, liu2020zero} and data augmentation based method \cite{chugh2019fingerprint, gajawada2019universal}.
\begin{figure*}
  \centering
  \includegraphics[width=.89\textwidth]{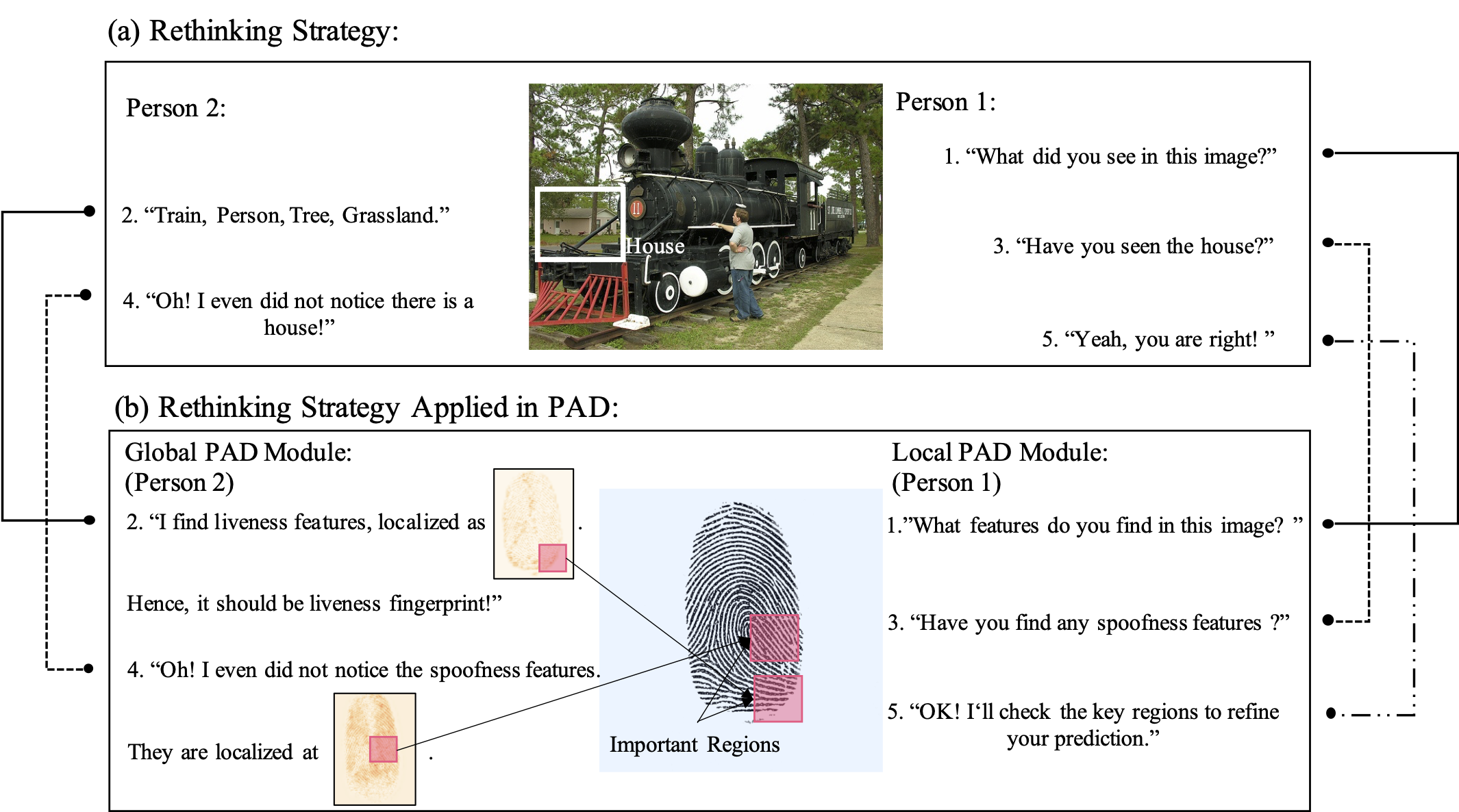}
  \caption{Samples of rethinking strategy: (a). Observation of rethinking strategy used in the vision system of human beings \cite{cao2015look}. (b). Our proposed scheme of rethinking strategy for fingerprint PAD. Person 1 in (a) corresponds to the local PAD module in (b), while Person 2 refers to the global PAD module.
  }
  \label{fig:rethink}
\end{figure*}

In the case of one-class-based methods, only bonafides (i.e., real fingerprints) are used to train the model. Once the concept of “bonafide” has been learned, spoofs of any material can be rejected.
For example, Engelsma et al. \cite{engelsma2019generalizing} proposed a Generative Adversarial Networks(GANs) based one-class classifier (1-class GANs).
The features learned by the discriminator in GANs to separate real live fingerprints from synthesized live fingerprints can be used during testing to distinguish live fingerprints from spoof fingerprints.
Liu et al. \cite{liu2020zero} proposed an auto-encoder-based method.
The reconstruction error and latent code obtained from the auto-encoder network are taken as the basis of the spoofness score for PAD.
However, both aforementioned methods are sensor-specific.
The method in \cite{engelsma2019generalizing} is based on Raspireader \cite{engelsma2018raspireader} and the sensor of the method given in \cite{liu2020zero} is OCT-based device \cite{liu2020flexible}.
Sensor-specific approaches rely on their expensive sensors and further limit the applications.
In order to solve this problem, a data augmentation-based method is proposed.
Typically, Chugh et al. \cite{chugh2019generalization} proposed a style-transfer-based wrapper, which transfers the style (texture) characteristics for data augmentation with the goal of improving the robustness against unknown materials.
Then the synthesized images are used to train the PA detector to improve the generalization performance.
However, this method needs the images from the target material or target sensor to obtain the style characteristics, which may not be accessible.
Different from the PAD method mentioned above, in this paper, a robust PAD method based on a rethinking strategy is proposed to improve the generalization performance.
Without using a specific sensor and the images from the target material or sensor, the proposed method can improve the performance of generalization effectively and outperform the state-of-the-art method following the settings used in LivDet 2017 \cite{mura2018livdet}, which is conveniently accessible in a real-world scenario.

\begin{figure*}
  \centering
  \includegraphics[width=.88\textwidth]{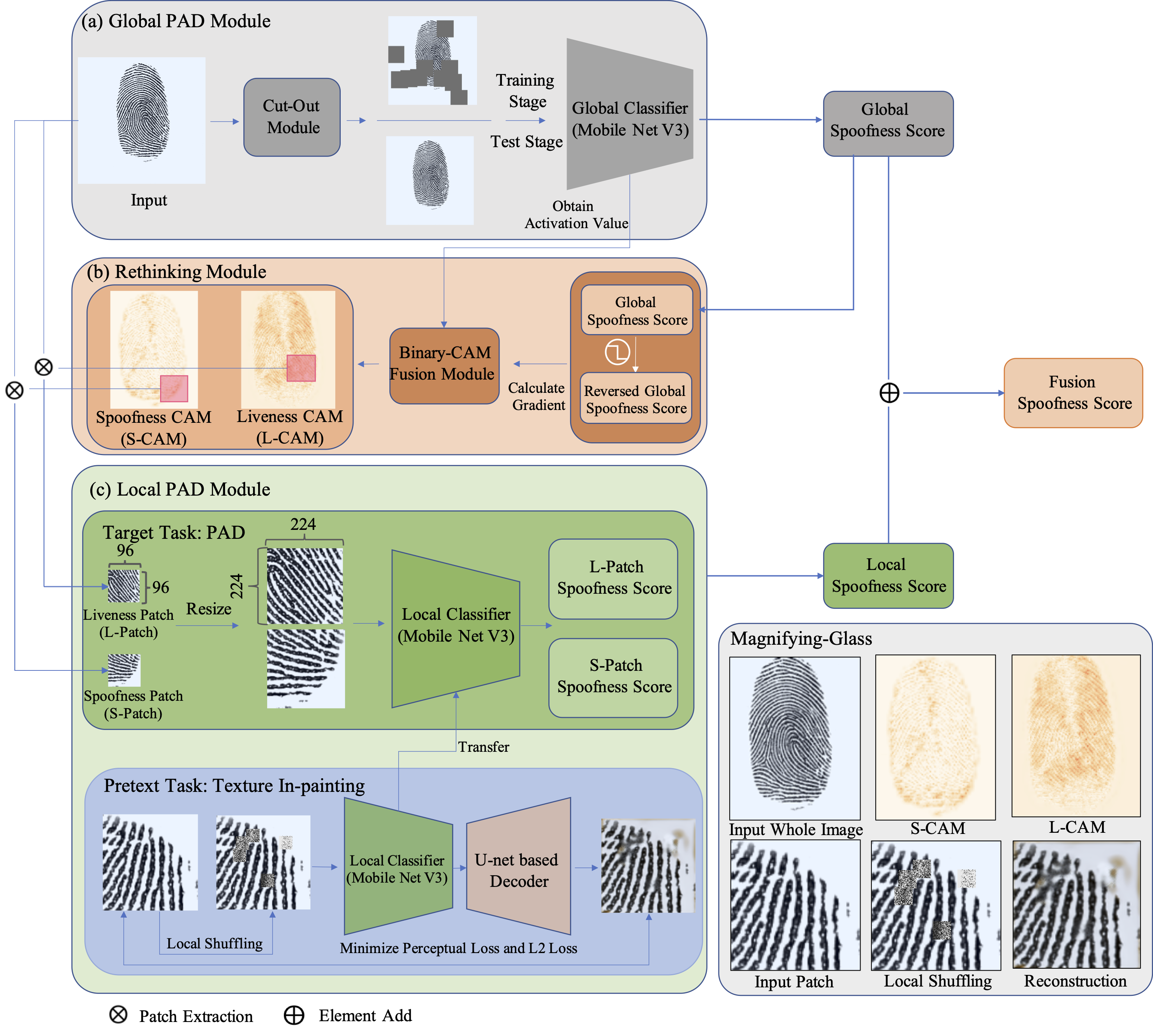}
  \caption{Flowchart of the proposed method, denoted as RTK-PAD. RTK-PAD consists of three modules: (a). Global PAD module, (b) Rethinking module, and (c) Local PAD module.
  }
  \label{fig:pipeline}
\end{figure*}

The proposed method in this paper is based on rethinking strategy.
As shown in Fig. \ref{fig:rethink} (a), rethinking strategy typically is a top-down manner dominated by “goals” from our mind \cite{beck2009top, cao2015look}.
By rethinking the location of class-specific image objects, human detection and recognition performances increase significantly.
This strategy is mainly used in two computer vision tasks, i.e., visual explanation from deep networks \cite{selvaraju2017grad, chattopadhay2018grad,zhou2016learning,zeiler2014visualizing} and weakly-supervised semantic segmentation (object detection) \cite{wang2020self,kolesnikov2016seed,wei2017object,hou2018self}.
Visual explanation is important for researchers to understand how and why deep network models work. Zeiler et al. \cite{zeiler2014visualizing} introduced a visualization technique, i.e. Deconvolution, which reveals the input stimuli that excite individual feature maps at any layers in the model.
However, the fine-grained visualizations produced by this method are not class-discriminative.
Zhou et al. \cite{zhou2016learning} proposed a Class Activation Mapping (CAM) method mimicking the above-mentioned rethinking strategy.
This approach modifies deep classification networks by replacing fully-connected layers with convolutional layers and global average pooling, thus achieving class-specific feature maps from feedforward-based activation value and backward-gradient-based weights.
Moreover, Selvaraju et al. \cite{selvaraju2017grad} proposed a Gradient-weighted Class Activation Mapping (Grad-CAM) method, which is a strict generalization of the CAM \cite{zhou2016learning}.
Unlike the model-specific method, i.e. CAM, this method can be applied to any CNN-based model.
In the task of weakly-supervised semantic segmentation (object detection), a group of advanced researchers uses image-level classification labels to train models and obtains semantic segmentation (object detection) using rethinking strategy \cite{wang2020self, kolesnikov2016seed, wang2017instance, xue2019diod}.
Specifically, most of them refine CAM generated from the classification network to approximate the segmentation mask (object bounding box).
Kolesnikov et al. \cite{kolesnikov2016seed} proposed three principles, i.e., seed, expand, and constrain, to refine CAMs.
Image-level erasing \cite{wei2017object, hou2018self} is a popular improved CAM-based segmentation method,
which erases the discriminative part of the input images guided by CAM to make networks learn class-specific features from other regions and expand activations.
Wang et al. \cite{wang2020self} proposed a self-supervised equivariant attention mechanism to discover additional supervision and improve the performance of CAM in weakly supervised segmentation tasks.
These studies show that rethinking strategy is crucial for computer vision and has already earned successes in different tasks. However, the utilization of the rethinking strategy in image classification tasks, especially in PAD, is rarely discussed.
In this paper, we employ the rethinking strategy into the PAD method to fuse the information from different types of input.

Fig. \ref{fig:pipeline} shows the flowchart of the proposed PAD method, which consists of a global PAD module, a rethinking module, and a local PAD module. Given an entire image as input, the global PAD module first predicts the global spoofness score, and then the rethinking module uses the output of the global module to obtain the important regions for PAD. Finally, we crop the patches corresponding to the important regions and employ the local PAD module to these patches to refine the prediction of the global PAD module.
As shown in Fig. \ref{fig:rethink}(b), the proposed pipeline can simulate the rethinking model in the human vision system shown in Fig. \ref{fig:rethink}(a). By following the class-specific hint, the rethinking module can localize the regions with significant class-specific features.
Based on the proposed rethinking strategy, the patches with both rich liveness and spoofness features are extracted and can be used to refine the final result, as shown in the fifth step in Fig. \ref{fig:rethink}(b).
Experimental results show that the proposed method can significantly improve the performance of PAD, and outperform the existing methods in both cross-material and cross-sensor settings by a wide margin.

The main contributions of this paper are summarized as follows.
\begin{itemize}
  \item A novel rethinking module for PAD is proposed in this paper. By analyzing the output of the global PAD module, the proposed rethinking module can localize the patches with the discriminative liveness/spoofness features. Given those patches output from the rethinking module as the inputs of the local PAD module, more accurate local spoofness scores can be achieved. Thus, the global PAD module and the local PAD module are well connected through the proposed rethinking module for the final effective PAD.

  \item As global and local modules are connected by rethinking modules to detect PAs, one of the primary tasks in the proposed method is to guide global and local modules to learn contrastive features. To achieve this target, two flexible pretext tasks, including Cut-out and texture in-painting, are proposed. The cut-out module, applied in the global PAD module, randomly obscures the local patches of the entire image to guide the classifier to learn non-local features. The texture in-painting task, which requires the module to in-paint the patches with pixel-shuffling, is proposed to lead the local PAD module to pay more attention to the local features.

  \item Experimental results on LivDet 2017 suggest that the proposed method outperforms the existing methods under cross-material and cross-sensor settings. In the cross-material setting, compared with the state-of-the-art method, a 45\% reduction of ACE and a 12.9\% rise in TDR@FDR=1\% are achieved for different sensors. In the cross-sensor setting, the proposed method improves the average PAD performance ACE from 32.40\% to 21.87\%.
\end{itemize}

\section{Presentation Attack Detection using RTK-PAD}
\label{sec:method}
As shown in Fig. \ref{fig:pipeline}, our method, denoted as RTK-PAD, applies a global-local model for PAD. We present the method from three parts, i.e. global PAD module, rethinking module, and local PAD module.
In global PAD module, a deep CNN based classifier using global image as input (i.e., global classifier shown in Fig. \ref{fig:pipeline}) is employed to obtain the feedforward-based activation value and global spoofness score.
\begin{algorithm}[!htb]
  \label{algo:1}
	\caption{Global-Local Model based PAD Method (RTK-PAD), including Training and Test Stage.}
	\begin{algorithmic}
    \Require\\
    Training Set of Images: $X_{train}=\{ x_1,...,x_i,...,x_n\}$;\\
    Groundtruth of $X_{train}$: $Y_{train}=\{ y_1,...,y_i,...y_n\}$;\\
    Test Sample : $x_t$; \\
    Global PAD Classifier: $GF( \cdot)$ with learning parameters $\theta_g$; \\
    Local PAD Classifier: $LF( \cdot )$ with learning parameters $\theta_l$; \\
    Cut-out Module : $CT(\cdot)$ \\
    Binary-CAM Fusion Module: BCFM($\cdot$);\\
    Loss Function: Classification Loss $\mathcal{L}_c(\cdot,\cdot)$\\
    Training Epoch Number: $e_g$ for $GF$ and $e_l$ for $LF$;\\
    Learning Rate: $\alpha_g$ for $GF$ and $\alpha_l$ for $LF$;
    \Ensure \State Fusion Spoofness Score;
  \end{algorithmic}
  \begin{algorithmic}[1]
    \State // \textbf{Training Stage}
    \State // \textbf{Training} $GF( \cdot )$ by minimizing $\sum_{i=1}^n L_c(GF(x_i),y_i)$
    \State Initialize $\theta_g$ from the model pretrained on ImageNet;
    \For{$i=1$ to $e_g$}
      \For{$X_{batch} \in  X_{train}$}
      \State $X_{batch} \leftarrow CT(X_{batch})$
      \State Gradient $\leftarrow \nabla_{\theta_g}\sum_{x_i \in X_{batch}}\mathcal{L}_c(GF(x_i),y_i)$;
      \State  Update $\theta_g \leftarrow \theta_g - \alpha_g * \text{Gradient}$;
      \EndFor
    \EndFor
    \State // \textbf{Training} $LF( \cdot )$ by minimizing $\sum_{i=1}^n L_c(LF(x_i),y_i)$
    \State Initialize $\theta_l$ from the pretext task: Texture In-painting;
    \For{$i=1$ to $e_l$}
    \For{$X_{batch} \in  X_{train}$}
    \State $\hat{P}_{batch} \leftarrow \text{Variance-PatchExtract}(X_{batch})$
    \State Gradient $\leftarrow \nabla_{\theta_g}\sum_{p_j \in \hat{P}_{batch}}\mathcal{L}_c(GF(p_j),y_j)$;
    \State // $y_j$ refers to the groundtruth of $p_j$
    \State  Update $\theta_g \leftarrow \theta_g - \alpha_g * \text{Gradient}$;
    \EndFor
    \EndFor
    \State // \textbf{Test stage}
    \State // \textbf{Global PAD Module:}
    \State Evaluate different scale feature maps $A_{FT}^t = \{ FT^t_1, FT^t_2, FT^t_3\}$ and the global spoofness score $gy_t^p$ with respect to $GF(x_t)$;
    \State // \textbf{Rethinking Module:}
    \State $gy_i^r$ $\leftarrow$ $1 - gy_t^p$; // Reverse Function
    \State $\text{L-CAM, S-CAM} \leftarrow \text{BCFM}(A^t_{FT}, gy_i^r, gy_t^p)$
    \State L-Patch $\leftarrow$ CAMPatchExtract($x_t$,L-CAM);
    \State S-Patch $\leftarrow$ CAMPatchExtract($x_t$,S-CAM);
    \State //\textbf{Local PAD Module:}
    \State $ly_t^l \leftarrow LF(\text{L-Patch})$; $ly_t^s \leftarrow LF(\text(S-Patch))$;
    \State // \textbf{Fusion Spoofness Score}
    \State $fy_t \leftarrow \frac{1}{3} (ly_i^l + ly_i^s + gy_i^{p})$
    \State Return $fy_t$
	\end{algorithmic}
\end{algorithm}
Given the input of activation value and spoofness score, the proposed rethinking module calculates the gradients of global classifier, and then obtains the spoofness CAM (S-CAM) and liveness CAM (L-CAM).
In local PAD module, L-CAM and S-CAM are used to crop two patches, called L-Patch and S-Patch, from the original input image. And a classifier (i.e., local classifier shown in Fig. \ref{fig:pipeline}) is applied to predict the local spoofness scores for both given patches.
Finally, the fusion spoofness score is calculated by averaging both local and global spoofness scores mentioned above.
In this section, we firstly illustrate on the system framework, then detail the Cut-out based global PAD module, local PAD module, rethinking module and the fusion spoofness score calculation.
\subsection{System Framework}
\label{sec:SF}
As shown in Fig. \ref{fig:pipeline}, the system mainly consists of three components, i.e., global PAD module, rethinking module and local PAD module.
While, global and local PAD modules are used to predict the spoofness score of the input fingerprint image, while rethinking part is used to obtain the important regions of input entire image and apply the corresponding patches to local PAD module.
As detailed in Algorithm. 1, we first trained global PAD classifier, $GF(\cdot)$, and local PAD classifier $LF(\cdot)$, which are the cores of the corresponding modules.
Given a test sample $x_t$, global PAD module obtained a set of different scale feature maps $A^t_{FT} = \{ FT^t_1, FT^t_2, FT^t_3\}$ and the global spoofness score $gy_t^{p}$ from $x_t$.
Then, on the basis of $gy_t^p$, rethinking module got the reversed global spoofness score $gy_t^{r}$ and applied binary-CAM fusion module to obtain the L-CAM and S-CAM.
Subsequently, we cropped $x_t$ into L-Patch and S-Patch with respect to L-CAM and S-CAM separately.
Given L-Patch and S-Patch as input, the local spoofness scores, $ly_t^l$ and $ly_t^s$ are obtained by local PAD module and, finally, the fusion spoofness score $fy_t$ can be calculated by
\begin{align}
  fy_t = \frac{1}{3} (ly_t^l + ly_t^s + gy_t^{p}) \label{eq:fusion}
\end{align}
\subsection{Cut-out based Global PAD Module}
The first step of RTK-PAD method, as given in Fig. \ref{fig:pipeline}, is to train the global classifier, $GF(\cdot)$.
As global and local PAD modules are fused for PAD, the learned features of each modules should be contrastive. Hence, Cut-out module, $CT(\cdot)$, is applied to train the classifier, leading $GF(\cdot)$ to learn non-local features, which is different from local PAD module.
Cut-out module is an extension of drop out in input space, i.e., a data regularization method. In training stage, we apply a fixed-size zero-mask to a random location of each input image. Specifically, inspired by \cite{devries2017improved}, the square patch is used as the Cut-out region. When Cut-out is applied to the fingerprint image, we randomly select a fixed-size path placing with a zero-mask. The training stage of $GF(\cdot)$ is given by
\begin{align}
 \min_{\theta_g} & \sum_{\substack{ x_i \in X_{train} \\ y_i \in Y_{train}}} \mathcal{L}_c(GF(CT(x_i)), y_i) \\
 \mathcal{L}_c(gy_i^p, y_i) &= y_i log(gy_i^p) + (1-y_i) log(1-gy_i^p)
\end{align}
where $gy_i^p$ refers to  $GF(CT(x_i))$.
In test stage, Cut-out module do not change the input images. Hence, the global spoofness score of $x_t$ is calculated as:
\begin{align}
  gy_t^p = GF(x_t) \label{eq:global}
\end{align}
For $GF(\cdot)$, the Mobile Net V3 architecture \cite{howard2019searching} is used as the backbone. Compared with the other popular deep networks, Mobile Net V3 is a low-latency architecture with the small number of model parameters, which is proved to be suitable for fingerprint PAD in \cite{chugh2018fingerprint}. On the basis of network architecture search (NAS), Mobile Net V3 uses a combination of depth-wise separable convolution \cite{howard2017mobilenets} , linear bottleneck \cite{sandler2018mobilenetv2}, inverted residual structure \cite{sandler2018mobilenetv2} and squeeze and excitation module \cite{tan2019mnasnet} as building blocks. In this redesigned building block, denoted as the bottleneck, point-wise convolution is firstly used to expand the size of features in channel dimension, and then, depth-wise separable convolution layer is applied to further extract the discriminative characteristics, followed by a squeeze and excitation module to weight the feature maps. Finally, the weighted feature map is processed by another point-wise convolution to reduce the size of features, and combined with the input of the bottleneck using inverted residual structure.
As the specification listed in TABLE \ref{tab:architecture}, the version of Mobile Net V3 used in this paper is called Large-Mobile Net V3, which consists of 14 bottlenecks, 4 convolution layers and 1 average pooling layer.
In TABLE \ref{tab:architecture}, SE denotes whether there is a squeeze and excitation in that block. NL represents the type of nonlinearity used. RE refers to ReLU and HS denotes h-swish \cite{howard2019searching}. NBN denotes no batch normalization and bneck refers to the bottleneck mentioned above. As can be seen, some adaptive modifications are used on the basis of original Mobile Net V3 (Large). Since the sizes of fingerprint images derived from the various sensors are different, the average pooling layer (i.e. Avg Pool in TABLE \ref{tab:architecture}) is used to replace the conventional pooling layer with $7\times 7$ kernel. Through average pooling layer, the network can handle arbitrary sizes of input fingerprint images. On the other side, the last layer of the original architecture, a convolution layer with 1000-unit output (designed for the 1000-class challenge of ImageNet dataset), is replaced by the convolution layer with 2-unit output of PAD, i.e. live vs. spoof.
\begin{table}[tbp]
  \centering
  \caption{Specification for Global Classifier and Local Classifier: Mobile Net V3 Architecture (Large Version) \cite{howard2019searching}}
  \resizebox{.48\textwidth}{32.4mm}{
  \begin{threeparttable}
  \begin{tabular}{ccccccc} 
  \toprule
  Input & Operator & Exp-Size  & Output-Channel & SE & NL & Stride  \\
  \midrule
  224$^2$ $\times$ 3 & conv2d 3$\times$3  & - & 16 & - & HS & 2 \\
  112$^2$ $\times$ 16 & bneck, 3$\times$3 & 16 & 16 & - & RE & 1 \\
  112$^2$ $\times$ 16\tnote{*} & bneck, 3$\times$3 & 64 & 24 & - & RE & 2 \\
  56$^2$ $\times$ 24 & bneck, 3$\times$3 & 72 & 24 & - & RE & 1 \\
  56$^2$ $\times$ 24 \tnote{*}& bneck, 5$\times$5 & 72 & 40 & $\surd$ & RE &2 \\
  28$^2$ $\times$ 40 & bneck, 5$\times$5 & 120 & 40 & $\surd$ & RE &1 \\
  28$^2$ $\times$ 40 & bneck, 5$\times$5 & 120 & 40 & $\surd$ & RE &1 \\
  28$^2$ $\times$ 40 \tnote{*}& bneck, 3$\times$3 & 240 & 80 & - & HS & 2 \\
  14$^2$ $\times$ 80 & bneck, 3$\times$3 & 200 & 80 & - & HS & 1 \\
  14$^2$ $\times$ 80 & bneck, 3$\times$3 & 184 & 80 & - & HS & 1 \\
  14$^2$ $\times$ 80 & bneck, 3$\times$3 & 480 & 112 & $\surd$ & HS & 1 \\
  14$^2$ $\times$ 112 & bneck, 3$\times$3 & 672 & 112 & $\surd$ & HS & 1 \\
  14$^2$ $\times$ 112 & bneck, 5$\times$5 & 672 & 160 & $\surd$ & HS & 2 \\
  7$^2$ $\times$ 160\tnote{*} & bneck, 5$\times$5 & 960 & 160 & $\surd$ & HS & 1 \\
  7$^2$ $\times$ 160 & bneck, 5$\times$5 & 960 & 160 & $\surd$ & HS & 1 \\
  7$^2$ $\times$ 160 & conv2d 1$\times$1 & - & 960 & - & HS & 1 \\
  7$^2$ $\times$ 960 & Avg Pool & - & - & - & - & 1 \\
  1$^2$ $\times$ 960\tnote{*} & conv2d 1$\times$1, NBN & - & 1280 & - & HS & 1 \\
  1$^2$ $\times$ 1280 & conv2d 1$\times$1, NBN & - & 2 & - & - & 1 \\
    \bottomrule
  \end{tabular}
  \begin{tablenotes}
       \footnotesize
       \item[*] The set of feature map $ B_{FT} = \{ FT_{1}, FT_{2}, FT_{3}, FT_{4}, FT_{5} \}$, and subset $A_{FT} = \{ FT_{1}, FT_{2}, FT_{3} \}$. The index order is from top to bottom.
     \end{tablenotes}
\end{threeparttable}
}
  \label{tab:architecture}
\end{table}
\subsection{Texuture In-painting based Local PAD Module}
Since $GF(\cdot)$ is guided to extracted non-local features, local PAD module should guide the local classifier, $LF(\cdot)$ to learn local characteristics to complement the limitation caused by single-level features. To achieve the goal, we apply the same architecture used in global PAD module, i.e. Mobile Net V3, but different type of input and pretext task in local PAD module.
Specifically, the patches cropped from the region of interest (ROI) of original fingerprint images are given as the input and texture in-painting task is applied to pretrain the local classifier.

As can be seen, in order to obtain the patches training local classifier, the ROI of raw images should be calculated at first. In this paper, a variance based method is used to determine whether the given patch belongs to ROI or background.
Given an original fingerprint image $x_i \in X_{train}$ as input, $x_i$ is cropped into a set of patches $P_i = \{ p_1, ..., p_j,..., p_N\}$ . Then, we calculate the variance, $\sigma^2_j$ of each patch $p_j$ by
\begin{align}
  \sigma^2_j &= \frac{1}{H \times W} \sum_{\substack{0 < h < H \\ 0 < w < W }} (p_j(h,w) - \mu_j)^2 
\end{align}
\begin{align}
  \mu_j &= \frac{1}{H  \times W}\sum_{\substack{0 < h < H \\ 0 < w < W }} p_j(h,w)
\end{align}
where $\mu_j$, $H$ and $W$ refer to the mean, height and width of $p_j$ respectively. Since the patch with a fingerprint has more texture characteristics, it generally has larger variance. Based on this observation, the set of ROI patches, $\hat{P}_i$, of $x_i$ is obtained by
\begin{align}
\hat{P}_i  = \{p_j \;| \; p_j \in P_i \; ; \;  \sigma^2_j > t\}
\end{align}
where $t$ is a predefined threshold.
\begin{figure*}
  \centering
  \includegraphics[width=.89\textwidth]{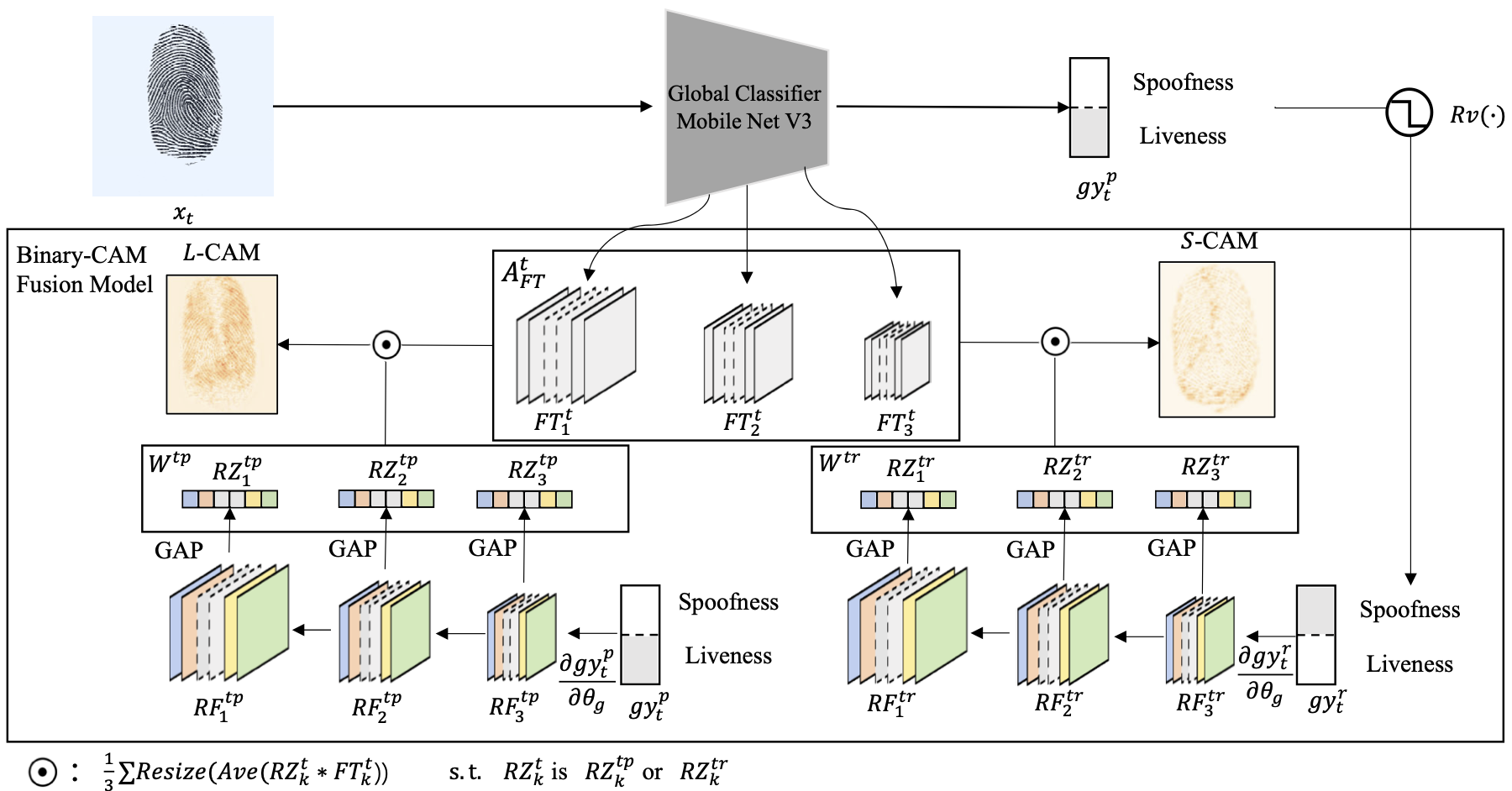}
  \caption{The flow chart of the rethinking module. While, the input of this module is different scale feature maps and the prediction of global PAD module, the output is L-CAM and S-CAM.
  }
  \label{fig:rethinking_model}
\end{figure*}

Based on the input $\hat{P}_i$, we then apply a pretext task, i.e. texture in-painting task, to pretrain $LF(\cdot)$, which is aimed to enhance the local feature extraction ability of $LF(\cdot)$.
As the pipeline shown in Fig. \ref{fig:pipeline}, given an original patch $p_j$, we sample some random windows from the patch followed by shuffling the order of contained pixels resulting in a transformed patch $p^s_j$.
The objective of this task is to in-paint $p^s_j$. Through $LF(\cdot)$, a set of different scale feature maps, $B_{FT}^{js} = \{ FT_1^{js}, ..., FT_k^{js}, ..., FT_5^{js}\}$ can be extracted from $p^s_j$. Among them, the specification of $FT_k$ is listed in TABLE \ref{tab:architecture}.
Since the ground truth of $p^s_j$ is $p_j$, a decoder, $DE(\cdot)$ is applied to combine with $B_{FT}^{js}$ in the U-net based way \cite{ronneberger2015u} to achieve patch in-painting. $DE(\cdot)$ and $LF(\cdot)$ can be trained end-to-end, and the training process can be given by,
\begin{align}
\min_{\theta_l,\theta_d} \mathcal{L}_{p}(D&E(B_{FT}^{js}), p_j) + \mathcal{L}_{l_2}(DE(B_{FT}^{js}), p_j) \label{eq:1}\\
\mathcal{L}_{p}(p^{re}_j,p_j) &= \frac{1}{3}\sum_{k=1}^{3} \mathcal{L}_{l2}(FT_k^{re},FT_k^j) \\
\mathcal{L}_{l_2}(p^{re}_j,p_j) &=  \frac{1}{H \times W}\!  \sum_{\substack{0 < h < H \\ 0 < w < W }} (p^{re}_j(h,w) - p_j(h,w))^2
\end{align}
where $\theta_l$ is the learning parameters of $LF(\cdot)$ and $\theta_d$ is the parameters of $DE(\cdot)$.  $p^{re}$ is the reconstruction result predicted by $DE(\cdot)$, i.e., $p^{re}_j = DE(B_{FT}^{js})$. $\mathcal{L}_{l_2}(\cdot,\cdot)$ represents the mean squared loss function and $\mathcal{L}_{p}(\cdot,\cdot)$ is the perceptual loss \cite{johnson2016perceptual} based on $LF(\cdot)$.
$FT_k^{re}$ and $FT_k^j$ are the $k$-th feature maps of $p^{re}_j$ and $p_j$ extracted by $LF(\cdot)$ respectively. Moreover, the order and location of $FT_k$ are illustrated in TABLE \ref{tab:architecture}.

On the basis of Eq. \eqref{eq:1}, we can obtain the trained $LF(\cdot)$ and then apply it to PAD. The training process can be written as:
\begin{align}
   \min_{\theta_l} & \sum_{\substack{ p_j \in \hat{P}_i \;;\; x_i \in X_{train} \\ y_i \in Y_{train}}} \mathcal{L}_c(LF(p_j), y_i)
\end{align}
Finally, the local spoofness score, $ly_t$ of the test patch, $p_t$ can be obtained by
\begin{align}
  ly_t = LF(p_t)
\end{align}

\subsection{Rethinking Module}

Since the features extracted by the local PAD module and the global PAD module are complementary, a fusion module should be applied to further improve the performance of PAD.
As shown in Fig. \ref{fig:rethink}, a rethinking module is proposed to fuse the information extracted by the local PAD module and the global PAD module. The rethinking module consists of reversing the prediction of the input fingerprint image and applying both the global predicted spoofness score and the reversed spoofness score into the binary-cam fusion module to localize the important region of the input image.

\begin{algorithm}[t]
  \label{algo:2}
	\caption{Binary-CAM Fusion Module}
	\begin{algorithmic}
    \Require\\
    Test Sample : $x_t$; \\
    Global PAD Classifier: $GF( \cdot)$ with learning parameters $\theta_g$; \\
    Global predicted spoofness score of $x_t$:$gy_t^p$\\
    Global reversed spoofness score of $x_t$:$gy_t^r$\\
    Assume that the prediction of $GF(x_t)$, i.e. $gy_t^p$, is liveness and $gy_i^r$ is spoofness.
    \Ensure \State L-CAM; S-CAM;
  \end{algorithmic}
  \begin{algorithmic}[1]
    \State $RF_k^{tp} \leftarrow \frac{\partial gy_t^p}{\partial FT_k}, \quad s.t. \quad k\in [1,3]$;
    \State $RF_k^{tr} \leftarrow \frac{\partial gy_t^r}{\partial FT_k}, \quad s.t. \quad k\in [1,3]$;
    \State $RZ_k^{tp} \leftarrow \frac{1}{H_k \times W_k} \sum_{0\!<\!h_k\!<\!H_k}\! \sum_{0\!<\!w_k\!<\!W_k} RF_k^{tp}(h_k,w_k,c_k)$;
    \State $RZ_k^{tr} \leftarrow \frac{1}{H_k \times W_k} \sum_{0\!<\!h_k\!<\!H_k}\! \sum_{0\!<\!w_k\!<\!W_k} RF_k^{tr}(h_k,w_k,c_k)$;
    \State $\text{L-CAM} \leftarrow \frac{1}{3}\sum_{k \in [1,3]} \text{resize}(\frac{1}{C_k}\sum_{0<c_k<C_k} RZ_k^{tp}(c_k) \times FT_k^t(h_k,w_k,c_k) )$
    \State $\text{S-CAM} \leftarrow \frac{1}{3}\sum_{k \in [1,3]} \text{resize}(\frac{1}{C_k}\sum_{0<c_k<C_k} RZ_k^{tr}(c_k) \times FT_k^t(h_k,w_k,c_k) )$
    \State Return L-CAM and S-CAM;
	\end{algorithmic}
\end{algorithm}

Fig. \ref{fig:rethinking_model} shows the pipeline of the proposed rethinking module. Given a test sample $x_t$, $A_{FT}^t$ and $gy_t^p$ are obtained from $GF(x_t)$. Then, a reverse function, $Rv(\cdot)$ is used to obtain the reversed global spoofness score, $gy_i^r$. $Rv(\cdot)$ is written as:
\begin{align}
  Rv(gy_t^p) = 1 - gy_t^p
\end{align}
Subsequently, $gy_t^p$ and $gy_i^r$ are applied to binary-cam fusion module to localize the import region of $x_t$. The overview of binary-cam fuision module is given in Algorithm. 2.
Since the importance of $\theta_g$ with respect to $gy_t^p$ (or $gy_i^r$) can be represented as $\frac{\partial gy_t^p}{\partial \theta_g}$ ($\frac{\partial gy_t^r}{\partial \theta_g}$), binary-cam fusion module first computes the gradient with respect to $FT_k$ to obtain the class-discriminative weight map $RF_k$. Considering PAD is a two-class problem, we applied both $gy_t^p$ and $gy_i^r$ to calculate $RF_k$, and this process can be written as:
\begin{align}
  RF_k^{tp} = \frac{\partial gy_t^p}{\partial FT_k} , \quad s.t. \quad k\in [1,3] \label{eq:2}
\end{align}
\vspace{-15pt}
\begin{align}
  RF_k^{tr} = \frac{\partial gy_t^r}{\partial FT_k} , \quad s.t. \quad k\in [1,3] \label{eq:3}
\end{align}
where $RF_k^{tp}$ is the weight map of $gy_t^p$ and $RF_k^{tr}$ refers to the weight map of $gy_t^r$. In Eq. \eqref{eq:2} and Eq. \eqref{eq:3}, we only compute the gradient of subset of $B_{FT}^t$, i.e. $FT_k \in A_{FT}^t $. This is due to the small size of $FT_4^t$ and $FT_5^t$, which can not provide valid information.
Considering $RF_k^{tp}$ (or $RF_k^{tr}$) $\in \mathbb{R}^{H_k \times W_k \times C_k}$, we can obatin a code $RZ_k^{tp}$ ($RZ_k^{tr}$) $\in \mathbb{R}^{C_k}$ by the operator of global average pooling(GAP). GAP can be written as:
\begin{align}
  RZ_k^{tp} =  \frac{1}{H_k \times W_k} \sum_{\substack{0 < h_k < H_k \\ 0 < w_k < W_k }} RF_k^{tp}(h_k,w_k,c_k)
\end{align}

\begin{align}
  RZ_k^{tr} =  \frac{1}{H_k \times W_k} \sum_{\substack{0 < h_k < H_k \\ 0 < w_k < W_k }} RF_k^{tr}(h_k,w_k,c_k)
\end{align}
where $c_k \in (0,C_k)$. In this paper, we assume that $gy_t^p$ indicates $x_t$ is liveness and $gy_t^r$ indicates $x_t$ is spoofness. Hence, $RZ_k^{tp}$ refers to the importance of each channel in $FT_k^t$ with respect to liveness.  On the other side, $RZ_k^{tr}$ represents the importance of spoofness. In order to obtain the liveness-discriminative localization map, denoted as L-CAM, and spoofness-discriminative localization map denoted as S-CAM, we fuse the different scale feature maps by
\begin{align}
  \text{L-CAM} \!\!= \!\!\frac{1}{3}\!\!\sum_{k\! \in\! [1,3]}\!\! \text{resize}(\frac{1}{C_k}\!\sum_{0\!<\!c_k\!<\!C_k}\!\!\! RZ_k^{tp}(c_k)\! \times \!FT_k^t(h_k,w_k,c_k))
\end{align}

\begin{align}
  \text{S-CAM} \!\!= \!\!\frac{1}{3}\!\!\sum_{k\! \in\! [1,3]}\!\! \text{resize}(\frac{1}{C_k}\!\sum_{0\!<\!c_k\!<\!C_k}\!\!\! RZ_k^{tr}(c_k)\! \times \!FT_k^t(h_k,w_k,c_k))
\end{align}
where resize$(\cdot)$ is an interpolation function, which converts the size of weighted feature maps into the size of $x_t$.
\subsection{Spoofness Score Calculation}
As L-CAM and S-CAM can be obtained through rethinking module, we can then localize the fixed-size patches of $x_t$ with the largest activation values in L-CAM or S-CAM. L-Patch refers to the patch with respect to L-CAM, and S-Patch corresponds to S-CAM.
Based on L-Patch and S-Patch, L-Patch spoofness score, $ly_t^l$ and S-Patch spoofness score, $ly_t^s$ can be computed by
\begin{align}
  ly_t^l = LF(\text{L-Patch})\\
  ly_t^s = LF(\text{S-Patch})
\end{align}
On the other side, we can obtain $gy_t^p$ by Eq. \eqref{eq:global}. Finally, as listed in Eq. \eqref{eq:fusion}, the average of $gy_t^p$, $ly_t^l$ and $ly_t^s$ is taken as the fusion spoofness score.
\begin{table*}[!htbp]
  \centering
  \caption{Number of samples for each sensor and subset of LivDet 2017 \cite{mura2018livdet}}
  \Huge
  \resizebox{.98\textwidth}{15.4mm}{
  \begin{tabular}{c|cccccccccccccc} 
  \toprule
  \multirow{3}{*}{Sensor} & \multicolumn{4}{c}{Training Set \Big{(}\tabincell{c}{ Images \\ Patches}\Big{)}} &~&\multicolumn{4}{c}{Validation Set  \Big{(}\tabincell{c}{ Images \\ Patches}\Big{)}} & ~ & \multicolumn{4}{c}{Test Set (Images)}\\
  \cline{2-5}
  \cline{7-10}
  \cline{12-15}
  & \multirow{2}{*}{Live} & \multicolumn{3}{c}{Spoofness Material} &~&\multirow{2}{*}{Live} & \multicolumn{3}{c}{Spoofness Material} & ~ & \multirow{2}{*}{Live} & \multicolumn{3}{c}{Spoofness Material} \\
  \cline{3-5}
  \cline{8-10}
  \cline{13-15}
  &  & Wood Glue & Ecoflex & Body Double &&~& Wood Glue & Ecoflex & Body Double &~&  & Gelatine & Latex & Liquid Ecoflex \\
  \midrule
    GreenBit & \tabincell{c}{800 \\ 800$\times$70=56000} & \tabincell{c}{320\\320$\times$70=22400} & \tabincell{c}{320\\320$\times$70=22400} & \tabincell{c}{320\\320$\times$70=22400}  & ~ & \tabincell{c}{200\\200$\times$70=14000} & \tabincell{c}{80\\80$\times$70=5600} & \tabincell{c}{80\\80$\times$70=5600} & \tabincell{c}{80\\80$\times$70=5600}  & ~ & 1700 & 680 & 680 & 680 \\
    \midrule
    Digital Persona & \tabincell{c}{799\\799$\times$70=55930} & \tabincell{c}{320\\320$\times$70=22400} & \tabincell{c}{320\\320$\times$70=22400} & \tabincell{c}{319\\319$\times$70=22330} & ~& \tabincell{c}{200\\200$\times$70=14000} & \tabincell{c}{80\\80$\times$70=5600} & \tabincell{c}{80\\80$\times$70=5600} & \tabincell{c}{80\\80$\times$70=5600}  & ~ & 1700 & 679 & 670 & 679 \\
    \midrule
    Orcanthus & \tabincell{c}{800 \\ 800$\times$70=56000} & \tabincell{c}{320\\320$\times$70=22400} & \tabincell{c}{320\\320$\times$70=22400} & \tabincell{c}{320\\320$\times$70=22400}  & ~ & \tabincell{c}{200\\200$\times$70=14000} & \tabincell{c}{80\\80$\times$70=5600} & \tabincell{c}{80\\80$\times$70=5600} & \tabincell{c}{80\\80$\times$70=5600}  & ~ &  1700 & 680 & 658 & 680 \\
    \bottomrule
  \end{tabular}
  }
  \label{tab:dataset}
\end{table*}
\begin{table*}[!htbp]
  \centering
  \Huge
  \caption{Performances of RTK-PAD with or without Each Proposed Module In Terms of ACE and TDR@FDR=1.0\%}
  \resizebox{.99\textwidth}{16.8mm}{
  \begin{tabular}{ccccccccccccccccc} 
  \toprule
  \multirow{2}{*}{Cut-out} & Global & Texture & Local & Rethinking &~&  \multicolumn{2}{c}{GreenBit} & ~& \multicolumn{2}{c}{DigitalPersona} &~& \multicolumn{2}{c}{Orcanthus}&~& \multicolumn{2}{c}{Mean $\pm$ s.d.} \\
  \cline{7-8}
  \cline{10-11}
  \cline{13-14}
  \cline{16-17}
  &PAD Module & In-painting & PAD Module & Module &~&ACE(\%) & TDR@FDR=1.0\%(\%)&~&ACE & TDR@FDR=1.0\%&~&ACE & TDR@FDR=1.0\%&~&ACE & TDR@FDR=1.0\%\\
  \midrule
  $\times$ &$\surd$ & $\times$ &$\times$ &$\times$ & ~ & 2.72 & 87.83 & ~ &4.12 & 77.47 & ~ & 3.74 & 83.30&~&3.53$\pm$0.59&82.87$\pm$4.24\\
  $\surd$ &$\surd$ & $\times$ &$\times$ &$\times$ & ~ & \textbf{2.20} & \textbf{94.70} & ~ & \textbf{3.31} & \textbf{80.47} & ~ & \textbf{1.85} & \textbf{95.49}&~&\textbf{2.45$\pm$0.62}&\textbf{90.22$\pm$6.90} \\
  \midrule
  $\times$ &$\times$ & $\times$ &$\surd$ &$\times$ & ~ & 3.84 & 67.53 & ~ & 5.77 & 71.65 & ~ & 6.06& 59.91&~&5.22$\pm$0.99&66.36$\pm$4.86\\
  $\times$ &$\times$ & $\surd$ &$\surd$ &$\times$ & ~ & \textbf{3.33} & \textbf{76.26} & ~ & \textbf{4.15} & \textbf{77.02} & ~ & \textbf{5.94} & \textbf{62.04} &~& \textbf{4.47$\pm$1.09}&\textbf{71.77$\pm$6.89} \\
  \midrule
  $\times$ &$\surd$ & $\surd$ &$\surd$ &$\surd$ & ~ & 3.11 & 84.60 & ~ & 4.10 & 73.96 & ~ & 4.29 & 68.53&~&3.83$\pm$0.52&75.70$\pm$6.67\\
  $\surd$ &$\surd$ & $\times$ &$\surd$ &$\surd$ & ~ & 2.00 & 96.31 & ~ & 3.39 & 72.19 & ~ & 1.76 & 95.34&~&2.38$\pm$0.72&87.95$\pm$11.15\\
  $\surd$ &$\surd$ & $\surd$ &$\surd$ &$\times$ & ~ & 1.99 & 94.95 & ~ & 3.26 & 75.84 & ~ & 2.02 & 94.55&~&2.42$\pm$0.59&88.45$\pm$8.92\\
  $\surd$ &$\surd$ & $\surd$ &$\surd$ &$\surd$ & ~ & \textbf{1.92} & \textbf{96.82} & ~ & \textbf{3.25} & \textbf{80.57} & ~ & \textbf{1.67} & \textbf{96.18}&~&\textbf{2.28$\pm$0.69}&\textbf{91.19$\pm$7.51}\\
    \bottomrule
  \end{tabular}
  }
  \label{tab:Ablation}
\end{table*}
\section{Experimental Results and Analysis}
\label{sec:experiment}
To evaluate the performance of the proposed method, we carried out extensive experiments on LivDet 2017 \cite{mura2018livdet}.  We first introduced the dataset and implementation details in subsection \ref{sec:dataset_implementation}. Subsection \ref{sec:validation} then proved  the effectiveness of the proposed RTK-PAD method. Finally, a comparison with current PAD methods on LivDet 2017 was shown in subsection \ref{sec:Comparsion}.
\subsection{Dataset and Implementation Details}
\label{sec:dataset_implementation}
The dataset used in the experiment is LivDet 2017. This dataset contains over 17,500 fingerprint images captured from three different scanners, i.e. Green Bit DactyScan84C, Orcanthus Certis2 Image and Digital Persona U.are.U 5160. In particular, Green Bit DactyScan84C is applied for Italian border controls and issurance of italian electronic documents. Orcathus Certis2 Image is a general sensor embedded in Personal Computer (PC). And Digital Persona U.are.U 5160 is used with mobile device (e.g. Nexus 7 tablet). The sensors in LivDet 2017 are suitable for different practical applications, the evaluation can thus effectively test the practical performances of AFRSs in the real world. As listed in TABLE \ref{tab:dataset}, for each sensor, around 1760 fingerprint images are used for training, 440 and about 3740 fingerprint images are used for validation and testing respectively. For local PAD module, 70 patches are cropped from each original fingerprint image for training and validation. Unlike other LivDet datasets, the materials used in the training set are different with that used in the test set. As a cross-material dataset, LivDet 2017 is the right benchmark to evaluate the generalization performance of PAD methods.

In the selection of hyperparameters in the method, we refer to the setting of \cite{chugh2019fingerprint,chugh2018fingerprint, devries2017improved, zhou2019models}. In Cut-out module, the number of windows is 10 and the size of windows is set to $96 \times 96$. For local classifier, we crop the patches with the size $96 \times 96$ from original fingerprint images and then resize the patches from $96 \times 96$ to $224 \times 224$ as the input size required by Mobile Net V3.  In texture in-painting task, the number of window is 5 and the patch size is $32 \times 32$.

Our implementation is based on the public platform pytorch \cite{paszke2017automatic}. We initialized the weights in each layer from the model pretrained in ImageNet. The Adam optimizer is used for our method, the learning rate of the Adam is set to 0.0001. $\beta_1$ and $\beta_2$ of Adam are set to 0.9 and 0.999, respectively.
Since fingerprints are presented in a single channel, the fingerprint images are duplicated channel-wise to expand the shape from 224$\times$224$\times$1 to 224$\times$224$\times$3.
Data augmentation techniques, including vertical flipping, horizontal flipping and random rotation are employed to ensure that the trained model is robust to the possible variations in fingerprint images.
Our work station's CPU is 2.8GHz, RAM is 32GB and GPU is NVIDIA TITAN RTX.

To evaluate the performance of the methods, Average Classification Error (ACE) and True Detection Rate(TDR)@False Detection Rate(FDR)=1\% are used. 
\begin{itemize}
    \item ACE reflects the classification performance, where the smaller error value indicates the better performance of the evaluated method.
    \item  TDR@FDR=1\% represents the percentage of PAs that can spoof the biometric system when the reject rate of normal users $\le$ 1\%. The larger value of TDR@FDR=1\% suggests better performance.
\end{itemize}
For comparison, the winner in LivDet 2017 \cite{mura2018livdet}, the state-of-the-art PAD method \cite{chugh2018fingerprint} and a GAN based data augmentation method \cite{chugh2019generalization} are employed in this paper.
\begin{figure*}
  \centering
  \includegraphics[width=.99\textwidth]{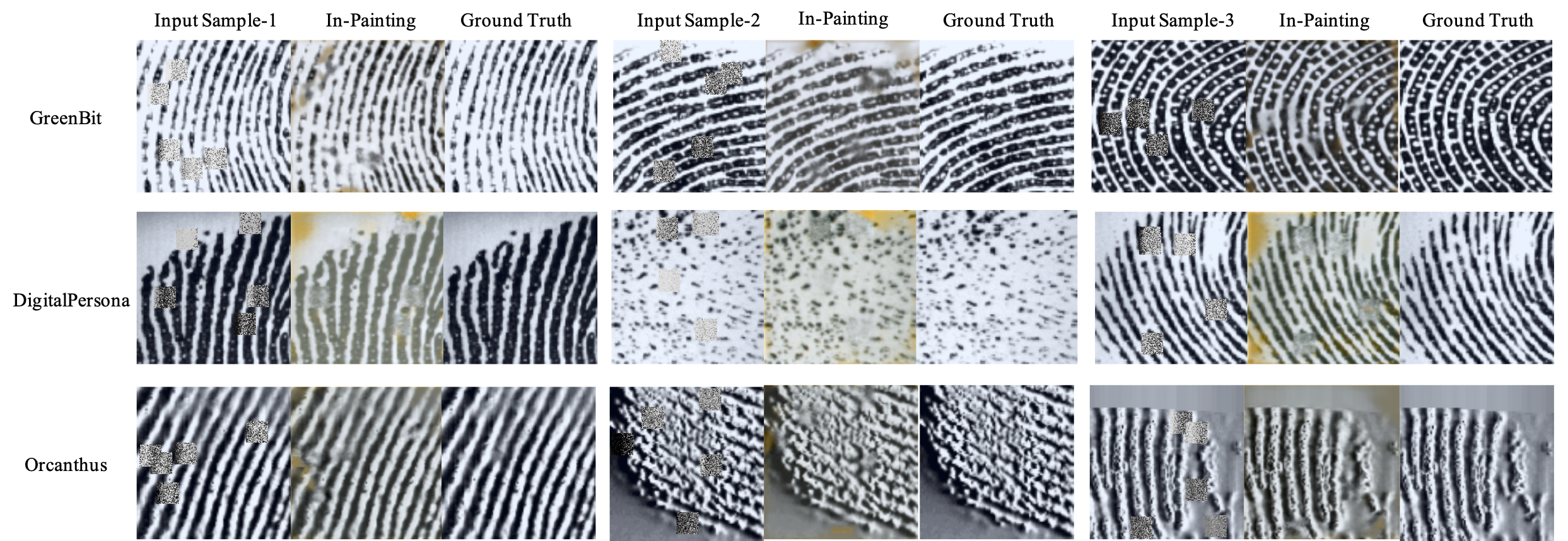}
  \caption{The in-painting samples, including inputs, in-painting images and ground truth. The first row shows the samples obtained from GreenBit. The second row refers to the samples derived from DigitalPersona. And the third row represents the samples captured by Orcanthus. In each block, the left image is the input, the middle one refers to the in-painting result and right one is the ground truth.
  }
  \label{fig:recon}
\end{figure*}
\begin{figure*}
  \centering
  \includegraphics[width=.99\textwidth]{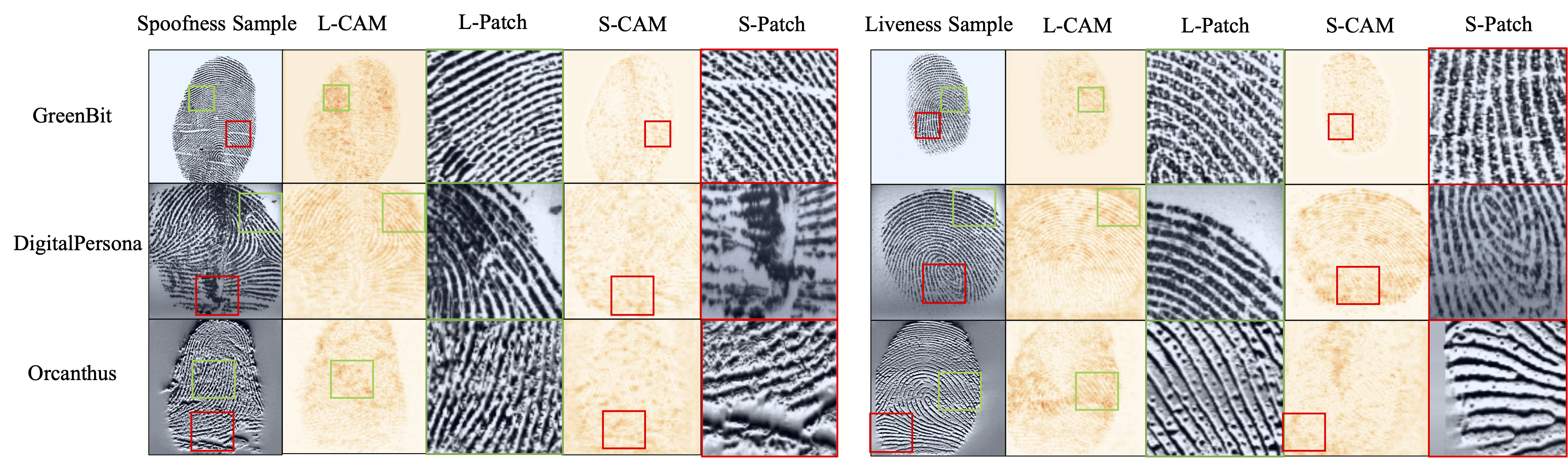}
  \caption{The samples of L-CAM, S-CAM, L-Patch and S-Patch. The first row shows the samples obtained from GreenBit. The second row refers to the samples derived from DigitalPersona. And the third row represents the samples captured by Orcanthus. In each line, the first block presents the L-CAM, L-Patch, S-CAM and S-Patch with respect to the spoofness sample, and the second one shows that of the liveness sample.
  }
  \label{fig:rethinking_visualization}
\end{figure*}
\begin{table*}[]
  \center
  \caption{Performance of RTK-PAD using Different Weighted Spoofness Scores In Terms of ACE and TDR@FDR=1.0\%}
    \resizebox{.98\textwidth}{15.2mm}{
\begin{threeparttable}
\begin{tabular}{cccccccccccc}
\toprule
Weights for & \multicolumn{2}{c}{GreenBit} & ~& \multicolumn{2}{c}{DigitalPersona} &~& \multicolumn{2}{c}{Orcanthus}&~& \multicolumn{2}{c}{ Mean $\pm$ s.d.} \\
\cline{2-3}
\cline{5-6}
\cline{8-9}
\cline{11-12}
 ($gy_t^p$,$ly_t^l$,$ly_t^s$) &ACE(\%) & TDR@FDR=1.0\%(\%)&~&ACE & TDR@FDR=1.0\%&~&ACE & TDR@FDR=1.0\%&~&ACE & TDR@FDR=1.0\%\\
\midrule
(1/4, 1/4, 1/2) & 2.20 & 94.24 &~& 4.03 & 72.33 &~& 4.41 & 89.25 &~& 3.55$\pm$0.96 & 85.27$\pm$9.38 \\
(1/4, 1/2, 1/4) & 2.71 & 92.68  &~& 3.88 & 71.30 &~& 4.17 & 91.33 &~& 3.59$\pm$0.63 & 85.10$\pm$9.78 \\
(1/2, 1/4, 1/4) & 2.08 & 96.26 &~& 3.25 & 78.50 & ~ & 1.61 & 96.63&~&2.31$\pm$0.69 & 90.46$\pm$8.46\\
(1/3, 1/3, 1/3) & 1.92 & \textbf{96.82} &~& 3.25 & \textbf{80.57} & ~ & 1.67 & 96.18&~&2.28$\pm$0.69&91.19$\pm$7.51 \\
\midrule
\multirow{2}{*}{\tabincell{c}{Grid Search$^*$ \\ Step = 0.1} \Big{(}\tabincell{c}{ Weights \\ Results}\Big{)}}
                & \multicolumn{2}{c}{(0.40, 0.30, 0.40) } &~&  \multicolumn{2}{c}{(0.10, 0.90, 0.20)  } & ~ & \multicolumn{2}{c}{(0.90, 0.50, 0.20)  }&~&\multirow{2}{*}{\textbf{2.12$\pm$0.72}}&\multirow{2}{*}{\textbf{91.20$\pm$7.59}} \\
                \cline{2-3}
                \cline{5-6}
                \cline{8-9}
                &  \textbf{1.78} & 96.41 &~& \textbf{3.12} & 80.47 &~& \textbf{1.46} & \textbf{96.73} &~&  & \\
\bottomrule
\end{tabular}
\begin{tablenotes}
       \small
       \item $^*$ In each case, weights are searched from 0 to 1 with 0.1 step, the evaluation metric is ACE and the accumulation of weights is not limited to 1.
\end{tablenotes}
\end{threeparttable}
}
\label{tab:weighting}
\end{table*}
\subsection{Effectiveness Validation of the Proposed RTK-PAD Method}
\label{sec:validation}
To quantify the contribution of each module in RTK-PAD, we tested discriminative performance of the method with or without each module. TABLE \ref{tab:Ablation} shows the results using three different scanners. From the results, it can be seen that, in global PAD module, all of the evaluated metrics achieved better performance when Cut-out module is employed for global PAD module. This indicates Cut-out module can improve the performance of global PAD module effectively.

In local PAD module, the performance of pretext task is tested. As shown in Fig. \ref{fig:recon}, the module can in-paint the ridge lines of the fingerprint effectively. This indicates the module can capture the direction and shape of the ridge lines, which are the typical local features. In order to further investigate the performance of the proposed pretext task, the baseline is set as the model pretrained from ImageNet. As listed in TABLE \ref{tab:Ablation}, the model pretrained from texture in-painting task outperforms the baseline significantly. Especially, an improvement of around 10\% in TDR@FDR=1\% can be achieved when texture in-painting is used as the pretext task. This proves that the feature space learned from nature images can not be directly applied to fingerprint images and the texture in-painting task proposed in this paper is useful for PAD model. In addition, the contribution of texture in-painting task using both local and global PAD modules is also tested carefully. As listed in  TABLE \ref{tab:Ablation}, texture in-painting task, combined with global PAD module, can improve the performance of PAD as well. Specifically, a 3.24\% rise in TDR@FDR=1.0\% can be achieved by texture in-painting task, which proves the complementarity between Cut-out and texture-inpainting tasks.

Since the rethinking module is the important part of the method, we also tested the contribution of rethinking module with respect to PAD as listed in TABLE \ref{tab:Ablation}. To ensure the fairness of the experiment, we define a fusion spoofness score using the entire image and two patches cropped from the ROI randomly as the baseline. Compared with the baseline, the method using rethinking module has a better performance.
\begin{table*}[!htbp]
  \centering
  \Huge
  \caption{Performance Comparison between the proposed method and state of the art results reported on LivDet2017 dataset for cross material experiments In Terms of ACE and TDR@FDR=1.0\%}
  \resizebox{.99\textwidth}{11.4mm}{
  \begin{tabular}{cccccccccccc} 
  \toprule
  \multirow{2}{*}{LivDet 2017} & ~ &  LivDet 2017 Winner\cite{mura2018livdet} & ~& \multicolumn{2}{c}{Fingerprint Spoof Buster \cite{chugh2018fingerprint}} &~& \multicolumn{2}{c}{Fingerprint Spoof Buster + UMG wrapper \cite{chugh2019generalization}} &~& \multicolumn{2}{c}{RTK-PAD} \\
  \cline{3-3}
  \cline{5-6}
  \cline{8-9}
  \cline{11-12}
   &~ & ACE(\%) & ~ &ACE(\%) & TDR@FDR=1.0\%(\%)&~&ACE & TDR@FDR=1.0\%&~&ACE & TDR@FDR=1.0\%\\
  \midrule
GreenBit &~& 3.56 &~& 3.32 & 91.07 &~&  2.58 & 92.29 &~& \textbf{1.92} & \textbf{96.82}\\
Digital Persona &~& 6.29 &~& 4.88 & 62.29 &~&  4.80 & 75.47 &~ & \textbf{3.25} & \textbf{80.57}\\
Orcanthus &~& 4.41 &~& 5.49 & 66.59 &~&  4.99 & 74.45 &~&\textbf{1.67} & \textbf{96.18}\\
Mean $\pm$ s.d.&~& 4.75 $\pm$ 1.40 &~& 4.56 $\pm$ 1.12& 73.32 $\pm$ 15.52 &~& 4.12 $\pm$ 1.34 & 80.74 $\pm$ 10.02 & ~ & \textbf{2.28} $\pm$ \textbf{0.69} & \textbf{91.19} $\pm$ \textbf{7.51}\\
    \bottomrule
  \end{tabular}
  }
  \label{tab:Comparison}
\end{table*}
\begin{table*}[!htbp]
  \centering
  \scriptsize
  \caption{Cross-Sensor Fingerprint Spoof Generalization Performance In Terms of ACE and TDR@FDR=1.0\%}
  \resizebox{.99\textwidth}{20.4mm}{
  \begin{tabular}{cccccccc} 
  \toprule
  \multicolumn{2}{c}{LivDet 2017} & ~ & \multicolumn{2}{c}{Fingerprint Spoof Buster \cite{chugh2018fingerprint}} &~& \multicolumn{2}{c}{RTK-PAD} \\
  \cline{1-2}
  \cline{4-5}
  \cline{7-8}
   Sensor in Training & Sensor in Testing & ~ &ACE(\%) & TDR@FDR=1.0\%(\%) &~&ACE & TDR@FDR=1.0\%\\
  \midrule
GreenBit & Orcanthus &~& 50.57 & 0.00  &~& \textbf{30.49} & \textbf{20.61}\\
GreenBit & DigitalPersona &~& 10.63 & 57.48  &~& \textbf{7.41} & \textbf{70.41}\\
Orcanthus & GreenBit &~& 30.07 & 8.02   &~& \textbf{28.81} & \textbf{15.00}\\
Orcanthus & DigitalPersona &~& 42.01 & 4.97  &~&  \textbf{29.19} & \textbf{13.26}\\
DigitalPersona & GreenBit &~& 10.46 & 57.06   &~& \textbf{6.74} & \textbf{70.25}\\
DigitalPersona & Orcanthus &~& 50.68 & 0.00   &~& \textbf{28.55} & \textbf{18.68}\\
\multicolumn{2}{c}{Mean $\pm$ s.d.} &~& 32.40$\pm$ 16.92 & 21.26 $\pm$ 28.06 &~& \textbf{21.87} $\pm$ \textbf{10.48} & \textbf{34.70} $\pm$  \textbf{25.30}\\
    \bottomrule
  \end{tabular}
  }
  \label{tab:cross_sensor}
\end{table*}
In the case of DigitalPersona, TDR@FDR=1.0\% of the baseline is 75.84\%, but the global classifier can achieve 80.47\% TDR@FDR=1.0\% without any fusion operation. This indicates simple fusion can not improve the performance effectively. For the model using rethinking module, ACE(\%) is decreased from 3.31 to 3.25, and TDR(\%)@FDR=1.0\% is improved from 80.47 to 80.57, surpassing the baseline significantly (80.57\% vs. 75.84\% in terms of TDR@FDR=1.0\%). On the other side, as shown in Fig. \ref{fig:rethinking_visualization}, the results of rethinking module were visualized. It can be seen, whether the input is spoofness or liveness, L-CAM and S-CAM have different distribution. The region with large values in L-CAM generally has small values in S-CAM and vice versa. This indicates L-Patch and S-Patch have quite different localization. Hence, the features given by L-Patch and S-Patch are representative for PAD. The visualization of L-Patch and S-Patch further proves this observation.
As the L-Patch and S-Patch shown in Fig. \ref{fig:rethinking_visualization}, given different class of the fingerprint images, the features of L-Patch and S-Patch are different. When a spoofness sample is given to the rethinking module, rethinking module prefers to localize the region with a clear ridge line as L-Patch, and the region with distortion and destruction as S-Patch. When the given sample is liveness, rethinking module always crops the regions with more pores as L-Patch, and the region with nonsequence as S-Patch. Among them, the pore is a distinct liveness feature and the fusion of distortion and destruction is a general spoofness feature. Consequently, the results of L-Patch and S-Patch are in accordance with prior knowledge, which prove the effectiveness of the rethinking module.

In this paper, the averaged weights (1/3, 1/3 and 1/3) are used for the fusion of one global and two local spoofness scores. In fact, we've tried other weights, such as (1/2, 1/4, 1/4) and grid searching with step of 0.1. The results are listed in TABEL \ref{tab:weighting}. As can be seen,  the fusion spoofness score based on the averaged weights achieves 2.28\% mean ACE and 91.19\% TDR@FDR=1.0\%, which are comparable with the best results (2.12\% mean ACE and 91.2\% TDR@FDR=1.0\%) obtained by grid searching. However, the time efficiency of grid searching is low, and the optimal weights for spoofness scores might be different in each case. We thus adopted averaged weights in the rest of paper for simplicity.

\subsection{Comparison with Existing Methods}
\label{sec:Comparsion}
To further verify the effectiveness of the proposed method, we compared it with three PAD methods introduced in related work, i.e., the winner in LivDet 2017 \cite{mura2018livdet}, the state-of-the-art PAD method \cite{chugh2018fingerprint} and a GAN based data augmentation method \cite{chugh2019generalization}. In this paper, two different experimental settings are considered, including cross-material and cross-sensor. In cross-material, as the materials listed in TABLE \ref{tab:dataset}, the spoof materials available in the test set are 'unknown' materials, since these are different from the materials included in the training set. Hence, the data partition of LivDet2017 can be directly used as the cross-material setting. In cross-sensor setting, the PAD models are trained in the training set of source sensor, say GreenBit, and then tested in the test set of the target sensor, say Orcanthus.

1) \textit{Cross-Material Fingerprint Spoof Generalization}

From TABLE \ref{tab:Comparison}, we can see that the proposed method outperforms all the competing methods by a wide margin. Compared with the combination of the state-of-the-art PAD classifier, Fingerprint Spoof Buster \cite{chugh2018fingerprint}, and data augmentation method, UMG wrapper \cite{chugh2019generalization}, a 45\% reduction in the mean of ACE and a 12.9\% rise in the mean of TDR@FDR=1\% are achieved by the proposed method. The most significant improvement caused by the method is in the case of Orcanthus. Given Orcanthus as the reader, the first place before is the winner in LivDet 2017 \cite{mura2018livdet} with 4.41\% ACE.
Our proposed method can achieve 1.67\% ACE, which outperforms the previous winner markedly. This proves that the proposed method has better performance and generalization ability in PAD.

2) \textit{Cross-Sensor Fingerprint Spoof Generalization}

To further show the robustness of the proposed method, we compared the performance of the proposed method with Fingerprint Spoof Buster \cite{chugh2018fingerprint} under cross-sensor setting. As the results shown in TABLE \ref{tab:cross_sensor}, the proposed method obviously outperforms Fingerprint Spoof Buster in every case. Typically, when the sensor in training is GreenBit and the sensor in testing is Orcanthus, the proposed method achieves 30.49\% in terms of ACE, exceeding Fingerprint Spoof Buster by 20\% absolutely. It should be noted that, GreenBit and DigitalPersona are the optical sensors. Orcanthus is based on thermal swipe. Hence, the performance of cross-sensor between Orcanthus and other sensors are lower than that of the other cases.

UMG wrapper is a GAN based data augmentation method, which can also solve the cross-sensor problem effectively, but this method should use the data in target sensor. UMG wrapper can achieve 19.37\% mean ACE using 3100 local patches from 100 live fingerprint images from target sensor. Without using these 100 images, the proposed method can achieve 21.87\% mean ACE, which is very similar to UMG wrapper. This indicates, besides data synthesizing, the proposed method can also improve the cross-sensor performance effectively, but does not depend on any additional data derived from target sensor.

\subsection{Computational Complexity Analysis}
\label{sec:Complexity}
\begin{table}[]
  \caption{Computational Complexity in Terms of MACs, Params and Time Cost.}
  \Huge
    \resizebox{.5\textwidth}{13.3mm}{
\begin{tabular}{cccc}
\toprule
\multicolumn{1}{c}{}    & \tabincell{c}{Input Size} & \tabincell{c}{FLOPs\\(GMACs)} & \tabincell{c}{Time Cost\\ (ms)} \\
\midrule
Global PAD Module &1$\times$224$\times$224$\times$3 & 0.23 & 13.9\\
Local PAD Module &2$\times$224$\times$224$\times$3  & 2$\times$0.23  & 11.4\\
Rethinking Module & - & - &212.0\\
\midrule
Fingerprint Spoof Buster \cite{chugh2018fingerprint} &  $\sim$48$\times$224$\times$224$\times$3        & $\sim$27.84(48$\times$0.58)         &      268.2                                     \\
RTK-PAD (ours)           &    1$\times$224$\times$224$\times$3      & \textbf{0.69(3$\times$0.23)}         &    \textbf{237.3}                                       \\ \bottomrule
\end{tabular}
}
\label{tab:complexity}
\end{table}
Here, we calculated the FLOPs (floating point operations) \cite{molchanov2016pruning} and Time Cost to analyze our method's computational complexity. As listed in TABLE \ref{tab:complexity}, both Time Cost and FLOPs of the proposed method are less than that of the existing method, Fingerprint Spoof Buster \cite{chugh2018fingerprint}.
While 48 minutiae-centered patches need to be passed through Fingerprint Spoof Buster to calculate the spoofness score, only 3 passes of patches are required by the proposed method.
Hence, a 30.9 ms reduction in terms of Time Cost can be achieved by RTK-PAD. As Fingerprint Spoof Buster can be deployed in mobile device \cite{chugh2019fingerprint}, RTK-PAD with lower computation can well meet the requirements of practical applications.
It should be noted that, the input of global PAD module is loaded from local disks, while the input of local PAD is read from memory (DRAM) following the pipeline of RTK-PAD. Hence, although FLOPs of local PAD Module are larger than global PAD module, less time is required by local PAD module.
\section{Conclusion}
\label{sec:conclusion}
To improve the PAD performance under cross-material and cross-sensor settings, this paper proposed a global-local model based method, i.e. RTK-PAD. The method has three components, including global PAD module, local PAD module and rethinking module. Global PAD module is applied to predict the  global spoofness score of the entire fingerprint images, and local PAD module is employed to predict the local spoofness score of the patches of fingerprints. Rethinking module is the core of RTK-PAD. On the basis of global PAD module, rethinking module localizes the important region of input image, which is representative and discriminative for PAD. Then, local PAD uses the patches from the important region as the input, and predicts the local spoofness scores. Finally, RTK-PAD averages the global spoofness score and local spoofness scores as the final fusion spoofness score for PAD. The effectiveness of the proposed method was proved by experiments on LivDet 2017. Compared with the state-of-the-art PAD methods, a 45\% reduction in the mean of ACE and a 12.9\% rise in the mean of TDR@FDR=1.0\% were achieved by our method.
\section*{Acknowledgment}
Haozhe Liu would like to thank the support, patience, and care from Yong Huang over the passed years.
\ifCLASSOPTIONcaptionsoff
  \newpage
\fi
\bibliographystyle{IEEEtran}
\bibliography{myreference}

\end{document}